\documentclass{article}

\usepackage{times}
\usepackage{epsfig}
\usepackage{graphicx}
\usepackage{amsmath}
\usepackage{amssymb}
\usepackage{bm}
\usepackage[hyphens]{url}
\usepackage{graphbox}
\usepackage{gensymb}
\usepackage{float}
\usepackage{multirow}
\usepackage{booktabs}
\usepackage[group-separator={,}]{siunitx}

\usepackage{hyperref}

\usepackage[accepted]{icml2018}

\usepackage{ifthen}
\usepackage{array}
\usepackage{xcolor}

\newcommand*{\todo}[2][]{\textcolor{red}{[\textbf{\ifthenelse{\equal{#1}{}}{TODO}{TODO(#1)}}: #2]}}

\newcolumntype{L}[1]{>{\raggedright\let\newline\\\arraybackslash\hspace{0pt}}m{#1}}
\newcolumntype{C}[1]{>{\centering\let\newline\\\arraybackslash\hspace{0pt}}m{#1}}
\newcolumntype{R}[1]{>{\raggedleft\let\newline\\\arraybackslash\hspace{0pt}}m{#1}}

\definecolor{legendgreen}{RGB}{39, 174, 96}
\definecolor{legendblack}{RGB}{0, 0, 0}
\definecolor{legendred}{RGB}{211, 47, 47}

\newcommand*\diff{\mathop{}\!\mathrm{d}}

\newcommand{\ql}{query-limited\ }

\begin{document}

%%%%%%%%% TITLE
\twocolumn[
\icmlsetsymbol{equal}{*}
\icmltitle{Black-box Adversarial Attacks with Limited Queries and Information}

\begin{icmlauthorlist}
    \icmlauthor{Andrew Ilyas}{equal,mit,ls}
    \icmlauthor{Logan Engstrom}{equal,mit,ls}
    \icmlauthor{Anish Athalye}{equal,mit,ls}
    \icmlauthor{Jessy Lin}{equal,mit,ls}
\end{icmlauthorlist}

\icmlaffiliation{mit}{Massachusetts Institute of Technology}
\icmlaffiliation{ls}{LabSix}

\icmlcorrespondingauthor{LabSix}{team@labsix.org}

\vskip 0.3in
]

\printAffiliationsAndNotice{\icmlEqualContribution} % otherwise use the standard text.

%%%%%%%%% ABSTRACT
\begin{abstract}
    Current neural network-based classifiers are susceptible to adversarial
examples even in the black-box setting, where the attacker only has
query access to the model. In practice, the threat model for real-world systems
is often more restrictive than the typical black-box model where the adversary
can observe the full output of the network on arbitrarily many chosen inputs.
We define three realistic threat models that more accurately characterize
many real-world classifiers: the query-limited setting, the
partial-information setting, and the label-only setting. We
develop new attacks that fool classifiers under these more restrictive threat
models, where previous methods would be impractical or
ineffective. We demonstrate that our methods are effective against an ImageNet
classifier under our proposed threat models.
We also
demonstrate a targeted black-box attack against a commercial classifier,
overcoming the challenges of limited query access, partial information, and
other practical issues to break the Google Cloud Vision API.

\end{abstract}

%%%%%%%%% BODY TEXT
\section{Introduction}
\label{sec:introduction}

%* why black-box matters
    %* give several examples, besides just self-driving car
%* focus on magnitude differences between ours and previous work, and talk about
  %how ours makes certain things feasible that basically were
  %intractable before
%* what is the threat model
%* what are the security parameters
    %* when #queries might be a security parameter (online attack)
    %* when it might not be: e.g. self-driving car
%* what are NOT security parameters
    %* computation time required by the adversary

Neural network-based image classifiers are susceptible to adversarial examples,
minutely perturbed inputs that fool
classifiers~\cite{szegedy,biggio2013evasion}. These adversarial examples can
potentially be exploited in the real world~\cite{goodfellow-physical,robustadv,sharif,stopsign}.
For many commercial or proprietary systems, adversarial examples must
be considered under a limited threat model. This has
motivated black-box attacks that do not require access to the gradient
of the classifier.

One approach to attacking a classifier in this setting trains a
substitute network to emulate the original network and then
attacks the substitute with first-order white-box
methods~\cite{papernot16,papernot17}. Recent works note that adversarial
examples for substitute networks do not always transfer to the target model,
especially when conducting targeted attacks~\cite{zoo,narodytska}. These works
instead construct adversarial examples by estimating the
gradient through the classifier with coordinate-wise finite difference methods.

We consider additional access and resource restrictions on the black-box model
that characterize restrictions in real-world systems. These restrictions render
targeted attacks with prior methods impractical or infeasible. We present new
algorithms for generating adversarial examples that render attacks in the
proposed settings tractable.

\subsection{Definitions}

At a high level, an adversarial example for a classifier is an input that is
slightly perturbed to cause misclassification.

Prior work considers various threat models~\cite{sp2016:papernot,sp2017-carlini}.
In this work, we consider $\ell_\infty$-bounded perturbation that causes
\textit{targeted} misclassification (i.e. misclassification as a given target
class). Thus, the task of the adversary is: given an input $x$, target class
$y_{adv}$, and perturbation bound $\epsilon$, find an input $x_{adv}$ such that
$||x_{adv} - x||_\infty < \epsilon$ and $x_{adv}$ is classified as $y_{adv}$.

All of the threat models considered in this work are additional restrictions on
the black-box setting:

\paragraph{Black-box setting. } In this paper, we use the definition of
    black-box access as query access~\cite{zoo,iclr2017-liu,hayes-universal}.
    In this model, the adversary can supply any input $x$ and receive
    the predicted class probabilities, $P(y | x)$ for all classes $y$.
    This setting does not allow the adversary to analytically compute the
    gradient $\nabla P(y|x)$ as is doable in the white-box case.

We introduce the following threat models as more limited variants of the
black-box setting that reflect access and resource restrictions in real-world
systems:

\begin{enumerate}
    \item \textbf{Query-limited setting. } In the \ql setting, the attacker
	has a limited number of queries to the classifier. In this setting,
	we are interested in query-efficient algorithms for generating
    adversarial examples. A limit on the number of queries can be a result of
    limits on other resources, such as a time limit if inference time is a
    bottleneck or a monetary limit if the attacker incurs a cost for each query.

        \underline{Example.} The Clarifai NSFW (Not Safe for Work) detection API\footnote{\url{https://clarifai.com/models/nsfw-image-recognition-model-e9576d86d2004ed1a38ba0cf39ecb4b1}} is a binary
    classifier that outputs $P(NSFW|x)$ for any image $x$ and can be queried
    through an API. However, after the first 2500 predictions, the Clarifai
    API costs upwards of \$2.40 per 1000 queries. This makes a 1-million
    query attack, for example, cost \$2400.

    \item \textbf{Partial-information setting. } In the
	partial-information setting, the attacker only has access
    to the probabilities $P(y|x)$ for $y$ in the top $k$ (e.g. $k = 5$) classes
    $\{y_1, \ldots, y_k \}$.
    Instead of a probability, the classifier may even output a
    score that does not sum to 1 across the classes to indicate
    relative confidence in the predictions.

    Note that in the special case of this setting where $k = 1$, the attacker
    only has access to the top label and its probability---a
    partial-information attack should succeed in this case as well.

    \underline{Example.} The Google Cloud Vision
        API\footnote{\url{https://cloud.google.com/vision/}} (GCV) only outputs
        scores for a number of the top classes (the number varies between
        queries). The score is not a probability but a ``confidence score''
        (that does not sum to one).

    \item \textbf{Label-only setting. } In the
	label-only setting, the adversary does
    not have access to class probabilities or scores.
	Instead, the adversary only has access to a list
    of $k$ inferred labels ordered by their predicted probabilities. Note
    that this is a generalization of the decision-only setting defined in \citet{brendel}, where
    $k=1$, and the attacker only has access to the top label. We aim to
    devise an attack that works in this special case but can exploit extra
    information in the case where $k > 1$.

    \underline{Example.} Photo tagging apps such as Google
    Photos\footnote{\url{https://photos.google.com/}} add labels to user-uploaded
    images. However, no ``scores'' are assigned to the labels, and so an
    attacker can only see whether or not the classifier has inferred a
    given label for the image (and where that label appears in the ordered list).
\end{enumerate}

\subsection{Contributions}

\paragraph{Query-efficient adversarial examples. } Previous methods using
substitute networks or coordinate-wise gradient estimation for targeted
black-box attacks require on the order of millions of queries to attack an
ImageNet classifier. Low throughput, high latency, and rate limits on
commercially deployed black-box classifiers heavily impact the feasibility of
current approaches to black-box attacks on real-world systems.

We propose the variant of NES described in \citet{salimans} (inspired by
\citet{nes}) as a method for generating adversarial examples in the
query-limited setting. We use NES as a black-box gradient estimation
technique and employ PGD (as used in white-box attacks) with the estimated
gradient to construct adversarial examples.

We relate NES in this special case with the finite difference method over
Gaussian bases, providing a theoretical comparison with previous attempts at
black-box adversarial examples. The method does not require a substitute
network and is 2-3 orders of magnitude more query-efficient than previous
methods based on gradient estimation such as \citet{zoo}. We show that our
approach reliably produces targeted adversarial examples in the black-box
setting.

\paragraph{Adversarial examples with partial information. } We present a new
algorithm for attacking neural networks in the partial-information setting. The
algorithm starts with an image of the target class and alternates between
blending in the original image and maximizing the likelihood of the target
class. We show that our method reliably produces targeted adversarial examples
in the partial-information setting, even when the attacker only sees
the top probability. To our knowledge, this is the
first attack algorithm proposed for this threat model.

We use our method to perform the
first targeted attack on the Google Cloud Vision API, demonstrating the
applicability of the attack on large, commercial systems: the GCV API
is an opaque (no published enumeration of labels), partial-information (queries
return only up to 10 classes with uninterpretable ``scores''),
several-thousand-way commercial classifier.

\paragraph{Adversarial examples with scoreless feedback. } Often, in deployed
machine learning systems, even the score is hidden from the attacker. We
introduce an approach for producing adversarial examples even when no scores of
any kind are available. We assume the adversary only receives the top $k$
sorted labels when performing a query. We integrate noise robustness as a proxy
for classification score into our partial-information attack to mount a targeted
attack in the label-only setting. We show that even in the decision-only
setting, where $k=1$, we can mount a successful attack.

\section{Approach}
\label{sec:approach}

We outline the key components of our approach for conducting an attack in each of the proposed
threat models. We begin with a description of our application of Natural
Evolutionary Strategies~\cite{nes} to enable query-efficient generation of
black-box adversarial examples. We then show the need for a new technique for
attack in the partial-information setting, and we discuss our algorithm for
such an attack. Finally, we describe our method for attacking a classifier with
access only to a sorted list of the top $k$ labels ($k \geq 1$). We have released full
source code for the attacks we describe~\footnote{\url{https://github.com/labsix/limited-blackbox-attacks}}.

We define some notation before introducing the approach.
The projection operator $\Pi_{[x-\epsilon, x+\epsilon]}(x')$
is the $\ell_\infty$ projection of $x'$ onto an $\epsilon$-ball around $x$.
When $x$ is clear from context, we abbreviate this as $\Pi_\epsilon(x')$,
and in pseudocode we denote this projection with the function
$\textsc{Clip}(x', x-\epsilon, x+\epsilon)$. We define the function
$\text{rank}(y|x)$ to be the smallest $k$ such that $y$ is in the top-$k$
classes in the classification of $x$. We use $\mathcal{N}$ and
$\mathcal{U}$ to represent the normal and uniform distributions
respectively.

\subsection{Query-Limited Setting}
In the query-limited setting, the attacker has a query budget $L$
and aims to cause targeted misclassification in $L$ queries or less.
To attack this setting, we can use ``standard'' first-order
techniques for generating adversarial
examples~\citet{iclr2015-goodfellow,sp2016:papernot,madry-adversarial,sp2017-carlini},
substituting the gradient of the loss function with an
estimate of the gradient, which is approximated by querying the classifier rather
than computed by autodifferentiation. This idea is used in~\citet{zoo},
where the gradient is estimated via pixel-by-pixel finite differences, and
then the CW attack~\cite{sp2017-carlini} is applied. In this section, we
detail our algorithm for efficiently estimating the gradient
from queries, based on the Natural Evolutionary Strategies approach
of~\citet{nes}, and then state how the estimated gradient is used to generate
adversarial examples.

\subsubsection{Natural Evolutionary Strategies}
\label{sec:nes}
To estimate the gradient, we use NES~\cite{nes}, a method for
derivative-free optimization based on the idea of a search
distribution $\pi(\theta|x)$. Rather than maximizing an objective
function $F(x)$ directly,
NES maximizes the expected value of the loss function under the
search distribution. This
allows for gradient estimation in far fewer queries than typical
finite-difference methods. For a loss function $F(\cdot)$ and a current
set of parameters $x$, we have from~\citet{nes}:
\begin{align*}
    \mathbb{E}_{\pi(\theta|x)}\left[F(\theta)\right] &= \int F(\theta)\pi(\theta|x) \diff \theta \\
    \nabla_x \mathbb{E}_{\pi(\theta|x)}\left[F(\theta)\right] &= \nabla_x \int F(\theta)\pi(\theta|x) \diff \theta \\
    &= \int F(\theta) \nabla_x \pi(\theta|x) \diff \theta \\
    &= \int F(\theta) \frac{\pi(\theta|x)}{\pi(\theta|x)} \nabla_x \pi(\theta|x) \diff \theta \\
    &= \int \pi(\theta|x) F(\theta) \nabla_x \log\left(\pi(\theta|x)\right) \diff \theta \\
    &= \mathbb{E}_{\pi(\theta|x)}\left[F(\theta)\nabla_x \log\left(\pi(\theta|x)\right)\right]
\end{align*}

In a manner similar to that in~\citet{nes}, we choose a search distribution
of random Gaussian noise around the current image $x$; that is, we have $\theta
= x + \sigma\delta$, where $\delta \sim \mathcal{N}(0, I)$. Like~\citet{salimans},
we employ antithetic sampling to generate a population of $\delta_i$ values: instead of
generating $n$ values $\delta_i\sim \mathcal{N}(0, I)$, we sample Gaussian noise for $i \in \{1, \ldots,
\frac{n}{2}\}$ and set $\delta_j = -\delta_{n-j+1}$ for $j \in
\{(\frac{n}{2}+1), \ldots, n\}$. This optimization has been empirically
shown to improve performance of NES. Evaluating the gradient with a population of $n$ points
sampled under this scheme yields the following variance-reduced gradient estimate:
$$\nabla\mathbb{E}[F(\theta)] \approx \frac{1}{\sigma n}\sum_{i=1}^n \delta_i
F(\theta + \sigma\delta_i)$$
Finally, we perform a projected gradient descent update \cite{madry-adversarial}
with momentum based on the NES gradient estimate.

The special case of NES that we have described here
can be seen as a finite-differences
estimate on a random Gaussian basis.

\citet{quasiortho} shows that for an $n$-dimensional space and $N$
randomly sampled Gaussian vectors $v_1 \ldots v_N$, we can lower bound the
probability that $N$ random Gaussians are $c$-orthogonal:
$$N\! \leq\! -e^\frac{c^2 n}{4}\ln\left(p\right)^\frac{1}{2}\!\!\implies\!\!
\mathbb{P}\left\{\frac{v_i\cdot v_j}{||v_i||||v_j||} \leq c\ \forall\ (i,
j)\right\} \geq p$$

Considering a matrix $\Theta$ with columns $\delta_i$, NES
gives the projection $\Theta (\nabla F)$, so we can use standard
results from concentration theory to analyze our estimate.
A more complex treatment is given in \citet{mixedgaussian}, but using a
straightforward application of the
Johnson-Lindenstrauss Theorem, we can upper and lower
bound the norm of our estimated gradient $\widehat{\nabla}$ in terms of the true
gradient $\nabla$. As $\sigma \rightarrow 0$, we have that:
$$\mathbb{P}\left\{(1\!-\!\delta)||\nabla||^2 \leq ||\widehat{\nabla}||^2 \leq
(1\!+\!\delta)||\nabla||^2\right\} \geq 1 - 2p$$
$$ \text{where } 0 < \delta < 1 \text{ and } N = O(-\delta^{-2}\log(p))$$
More rigorous analyses of these ``Gaussian-projected finite
difference'' gradient estimates and bounds \cite{nesterov} detail
the algorithm's interaction with dimensionality, scaling, and various other factors.

\subsubsection{Query-Limited Attack}
\begin{algorithm}[tb]
   \caption{NES Gradient Estimate}
   \label{alg:nes}
\begin{algorithmic}
    \STATE {\bfseries Input:} Classifier $P(y|x)$ for class $y$, image $x$
    \STATE {\bfseries Output:} Estimate of $\nabla P(y|x)$
    \STATE {\bfseries Parameters:} Search variance $\sigma$, number of
    samples $n$, image dimensionality $N$
   \STATE $g \gets \bm{0}_n$
   \FOR{$i=1$ {\bfseries to} $n$}
	\STATE $u_i \gets \mathcal{N}(\bm{0}_{N}, \bm{I}_{N\cdot N})$
	\STATE $g \gets g + P(y|x+\sigma\cdot u_i)\cdot u_i$
	\STATE $g \gets g - P(y|x-\sigma\cdot u_i)\cdot u_i$
   \ENDFOR
   \STATE {\bfseries return} $\frac{1}{2n\sigma} g$
\end{algorithmic}
\end{algorithm}

In the query-limited setting, we use NES as an unbiased, efficient gradient
estimator, the details of which are given in Algorithm~\ref{alg:nes}.
Projected gradient descent (PGD) is performed using the sign of the
estimated gradient:
$$x^{(t)} = \Pi_{[x_0-\epsilon, x_0 + \epsilon]}(x^{(t-1)} - \eta \cdot \text{sign}(g_t))$$
The algorithm takes hyperparameters $\eta$, the step size, and
$N$, the number
of samples to estimate each gradient. In the query-limited setting with a
query limit of $L$, we use $N$ queries to estimate each gradient and
perform $\frac{L}{N}$ steps of PGD.

\subsection{Partial-Information Setting}
\label{sec:partial-info-attack}

In the partial-information setting, rather than beginning with
the image $x$, we instead begin with an instance $x_0$ \textit{of the
target class} $y_{adv}$, so that $y_{adv}$ will initially appear in the top-$k$ classes.

At each step $t$, we then alternate between:

(1) projecting onto $\ell_\infty$ boxes
of decreasing sizes $\epsilon_t$ centered at the original image $x_0$, maintaining that the
adversarial class remains within the top-$k$ at all times:
\begin{align*}
    \epsilon_t &= \min \epsilon' \text{  s.t.
    $\text{rank}\left(y_{adv}|\Pi_{\epsilon'}(x^{(t-1)})\right) < k$}
\end{align*}

(2) perturbing the image to maximize the probability of the adversarial target class,
\begin{align*}
    x^{(t)} &= \arg\max_{x'} P(y_{adv}|\Pi_{\epsilon_{t-1}}(x'))
\end{align*}

We implement this iterated optimization using backtracking line search to find
$\epsilon_t$ that maintains the adversarial class within the top-$k$, and several iterations
of projected gradient descent (PGD) to find $x^{(t)}$. Pseudocode is shown in
Algorithm~\ref{alg:pi}. Details regarding further
optimizations (e.g. learning rate adjustment) can be found in our source code.

\begin{algorithm}[tb]
   \caption{Partial Information Attack}
   \label{alg:pi}
\begin{algorithmic}
    \STATE {\bfseries Input:} Initial image $x$, Target class $y_{adv}$,
    Classifier $P(y|x): \mathbb{R}^n\times \mathcal{Y} \rightarrow [0,
    1]^k$ (access to probabilities for $y$ \textbf{in top $k$}), image $x$
    \STATE {\bfseries Output:}  Adversarial image $x_{adv}$ with
    $||x_{adv} - x||_\infty \leq \epsilon$
    \STATE {\bfseries Parameters:} Perturbation bound $\epsilon_{adv}$,
starting perturbation $\epsilon_0$, NES Parameters ($\sigma, N, n$),
epsilon decay $\delta_\epsilon$, maximum learning rate $\eta_{max}$, minimum
learning rate $\eta_{min}$
    \STATE $\epsilon \gets \epsilon_0$
    \STATE $x_{adv} \gets$ image of target class $y_{adv}$
    \STATE $x_{adv} \gets \textsc{Clip}(x_{adv}, x - \epsilon, x + \epsilon)$
    \WHILE{$\epsilon > \epsilon_{adv} \text{ or } \max_{y} P(y|x) \neq y_{adv}$}
	\STATE $g \gets \textsc{NESEstGrad}(P(y_{adv}|x_{adv}))$
	\STATE $\eta \gets \eta_{max}$
	\STATE $\hat{x}_{adv} \gets x_{adv} - \eta g$
	\WHILE{\textbf{not }$y_{adv} \in \textsc{Top-k}(P(\cdot|\hat{x}_{adv}))$}
	    \IF{$\eta < \eta_{min}$}
		\STATE $\epsilon \gets \epsilon + \delta_\epsilon$
		\STATE $\delta_\epsilon \gets \delta_\epsilon/2$
		\STATE $\hat{x}_{adv} \gets x_{adv}$
		\STATE \textbf{break}
	    \ENDIF
	    \STATE $\eta \gets \frac{\eta}{2}$
	    \STATE $\hat{x}_{adv} \gets \textsc{Clip}(x_{adv} - \eta g, x-\epsilon, x+\epsilon)$
	\ENDWHILE
	\STATE $x_{adv} \gets \hat{x}_{adv}$
	\STATE $\epsilon \gets \epsilon - \delta_\epsilon$
    \ENDWHILE
    \STATE {\bfseries return} $x_{adv}$
\end{algorithmic}
\end{algorithm}

\subsection{Label-Only Setting}
Now, we consider the setting where we only assume access to the top-$k$
sorted labels. As previously mentioned, we explicitly include the setting
where $k=1$ but aim to design an algorithm that can incorporate extra
information when $k > 1$.

\begin{figure}[h]
    \begin{centering}
	\includegraphics[width=\linewidth]{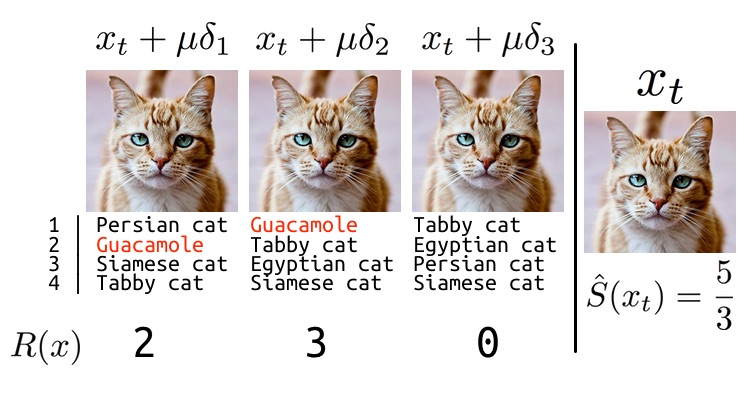}
	\caption{An illustration of the derivation of the proxy score
	$\hat{S}$ in the label-only setting.}
	\label{fig:labelonlyillustration}
    \end{centering}
\end{figure}

The key idea behind our attack is that in the absence of output
scores, we find an alternate way to
characterize the success of an adversarial example.
First, we define the \textit{discretized score} $R(x^{(t)})$ of an
adversarial example to quantify how adversarial the image is at each step $t$
simply based on the ranking of the adversarial label $y_{adv}$:
$$R(x^{(t)}) = k - \text{rank}(y_{adv}|x^{(t)})$$
As a proxy for the softmax probability, we consider the robustness of the
adversarial image to random perturbations (uniformly chosen from a $\ell_\infty$ ball of radius $\mu$), using the discretized
score to quantify adversariality:
$$S(x^{(t)}) = \mathbb{E}_{\delta \sim
\mathcal{U}[-\mu,\mu]}[R(x^{(t)} + \delta)]$$
We estimate this proxy score with a Monte Carlo approximation:
$$\widehat{S}(x^{(t)}) = \frac{1}{n}\sum_{i=1}^n R(x^{(t)} + \mu\delta_i)$$
A visual representation of this process is given in
Figure~\ref{fig:labelonlyillustration}.

We proceed to treat $\widehat{S}(x)$ as a proxy for the output probabilities $P(y_{adv} | x)$
and use the partial-information technique
we introduce in Section~\ref{sec:partial-info-attack} to find an adversarial example
using an estimate of the gradient $\nabla_x\widehat{S}(x)$.

\section{Evaluation}
\label{sec:evaluation}

We evaluate the methods proposed in Section~\ref{sec:approach} on their
effectiveness in producing targeted adversarial examples in the three threat
models we consider: query-limited, partial-information, and label-only. First,
we present our evaluation methodology. Then, we present evaluation results for
our three attacks. Finally, we demonstrate an attack against a commercial
system: the Google Cloud Vision (GCV) classifier.

\subsection{Methodology}

We evaluate the effectiveness of our attacks against an ImageNet classifier. We
use a pre-trained InceptionV3 network~\cite{szegedy-inception} that has 78\%
top-1 accuracy, and for each attack, we restrict our access to the classifier
according to the threat model we are considering.

For each evaluation, we randomly choose 1000 images from the ImageNet test set,
and we randomly choose a target class for each image. We limit $\ell_\infty$
perturbation to $\epsilon = 0.05$. We use a fixed set of hyperparameters across
all images for each attack algorithm, and we run the attack until we produce an
adversarial example or until we time out at a chosen query limit (e.g. $L =
10^6$ for the query-limited threat model).

We measure the \textit{success rate} of the attack, where an attack is
considered successful if the adversarial example is classified as the target
class and considered unsuccessful otherwise (whether it's classified as the
true class or any other incorrect class). This is a strictly harder task than
producing untargeted adversarial examples. We also measure the number of
queries required for each attack.

\subsection{Evaluation on ImageNet}

In our evaluation, we do not enforce a particular limit on the number of queries as
there might be in a real-world attack. Instead, we cap the number of
queries at a large number, measure the number of queries required for each
attack, and present the distribution of the number of queries required. For
both the the partial-information attack and the label-only attack, we
consider the special case where $k=1$, i.e. the attack only has access to
the top label. Note that in the partial-information attack the adversary also has
access to the probability score of the top label.

\begin{figure}
    \centering
    \includegraphics[width=\linewidth]{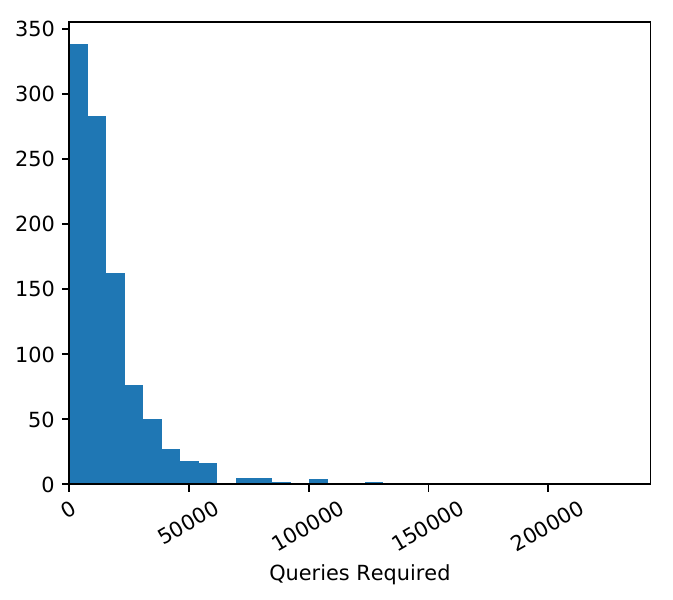}
    \includegraphics[width=\linewidth]{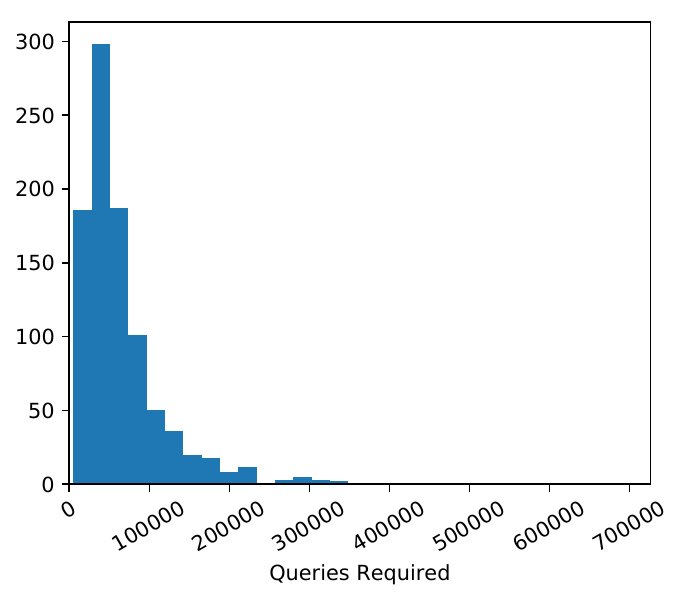}
    \caption{The distribution of the number of queries required for the
    query-limited (top) and partial-information with $k=1$ (bottom)
    attacks.}
    \label{fig:histograms}
\end{figure}

\begin{figure}
	\begin{centering}
        \includegraphics[width=\linewidth]{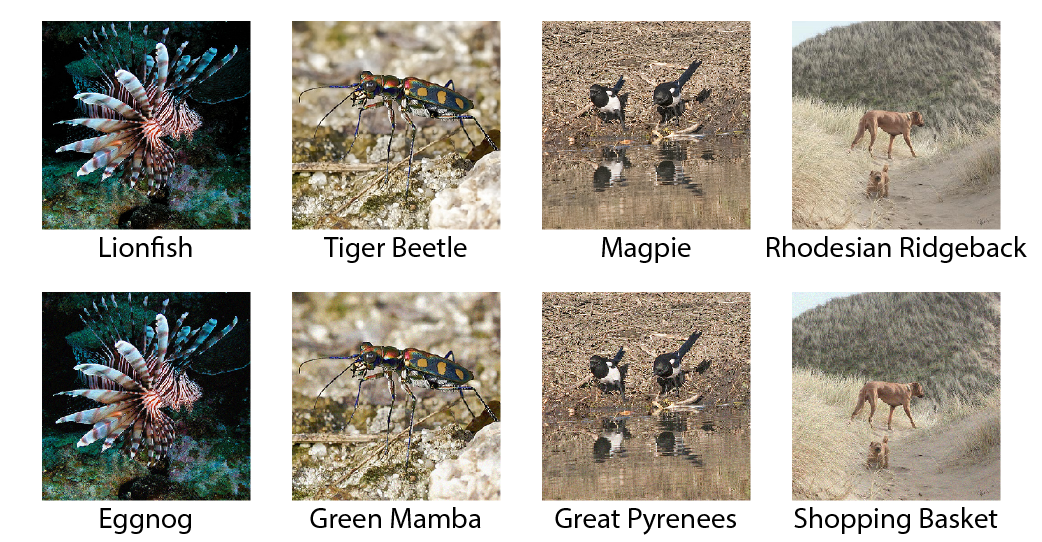}
        \caption{
            $\epsilon = 0.05$ targeted adversarial examples for the InceptionV3
            network. The top row contains unperturbed images, and the bottom
            row contains corresponding adversarial examples (with randomly
            chosen target classes).
        }
        \label{fig:imagenet-examples}
    \end{centering}
\end{figure}

Table~\ref{tab:results} summarizes evaluation results our attacks for the three
different threat models we consider, and Figure~\ref{fig:histograms} shows
the distribution of the number of queries.
Figure~\ref{fig:imagenet-examples} shows a sample of the adversarial examples
we produced.
Table~\ref{tab:params} gives our hyperparameters; for each attack, we use the
same set of hyperparameters across all images.

\begin{table}
    \centering
    \begin{tabular}{c c c c}
        \toprule
        \textbf{Threat model} & \textbf{Success rate} & \textbf{Median queries} \\
        \midrule
        QL & 99.2\% & \num{11550}
 \\
        PI & 93.6\% & \num{49624}
 \\
        LO & 90\% & \num{2.7e6}
 \\
        \bottomrule
    \end{tabular}
    \caption{
        Quantitative analysis of targeted $\epsilon = 0.05$ adversarial attacks
	in three different threat models: query-limited (QL),
	partial-information (PI), and label-only (LO). We perform attacks over 1000
	randomly chosen test images (100 for label-only) with randomly
	chosen target classes. For each attack, we use the same
	hyperparameters across all images. Here, we report the overall
	success rate (percentage of times the adversarial example was
	classified as the target class) and the median number of queries
	required.
    }
    \label{tab:results}
\end{table}

\begin{table}
\centering
\begin{tabular}{@{}ll@{}}
\toprule
\multicolumn{2}{l}{\textbf{General}} \\ \midrule
$\sigma$ for NES                                          & 0.001  \\
$n$, size of each NES population                          & 50   \\
$\epsilon$, $l_{\infty}$ distance to the original image & 0.05  \\
    $\eta$, learning rate                                             & 0.01  \\ \midrule
\multicolumn{2}{l}{\textbf{Partial-Information Attack}}	  \\ \midrule
$\epsilon_0$, initial distance from source image	  & 0.5 \\
$\delta_\epsilon$, rate at which to decay $\epsilon$      & 0.001 \\ \midrule
\multicolumn{2}{l}{\textbf{Label-Only Attack}}	  \\ \midrule
$m$, number of samples for proxy score			  & 50 \\
$\mu$, $\ell_\infty$ radius of sampling ball		  & 0.001 \\ \bottomrule
\end{tabular}
\caption{Hyperparameters used for evaluation}
\label{tab:params}
\end{table}

\subsection{Real-world attack on Google Cloud Vision}
\label{sec:gcv}

To demonstrate the relevance and applicability of our approach to
real-world systems, we attack the Google Cloud Vision (GCV) API, a publicly
available computer vision suite offered by Google. We attack the
most general object labeling classifier, which performs n-way
classification on images. Attacking GCV is considerably more challenging
than attacking a system in the typical black-box setting because of the
following properties:

\begin{itemize}
    \item The number of classes is large and
unknown --- a full enumeration of labels is unavailable.
    \item The classifier returns ``confidence scores'' for each label it assigns to an image,
which seem to be neither probabilities nor logits.
    \item The classifier does not return scores for all labels, but instead returns an
unspecified-length list of labels that varies based on image.
\end{itemize}

This closely mirrors our partial-information threat model, with the additional
challenges that a full list of classes is unavailable and the length of the
results is unspecified and varies based on the input. Despite these challenges,
we succeed in constructing targeted adversarial examples against this
classifier.

\begin{figure}
    \centering
    \includegraphics[width=0.8\linewidth]{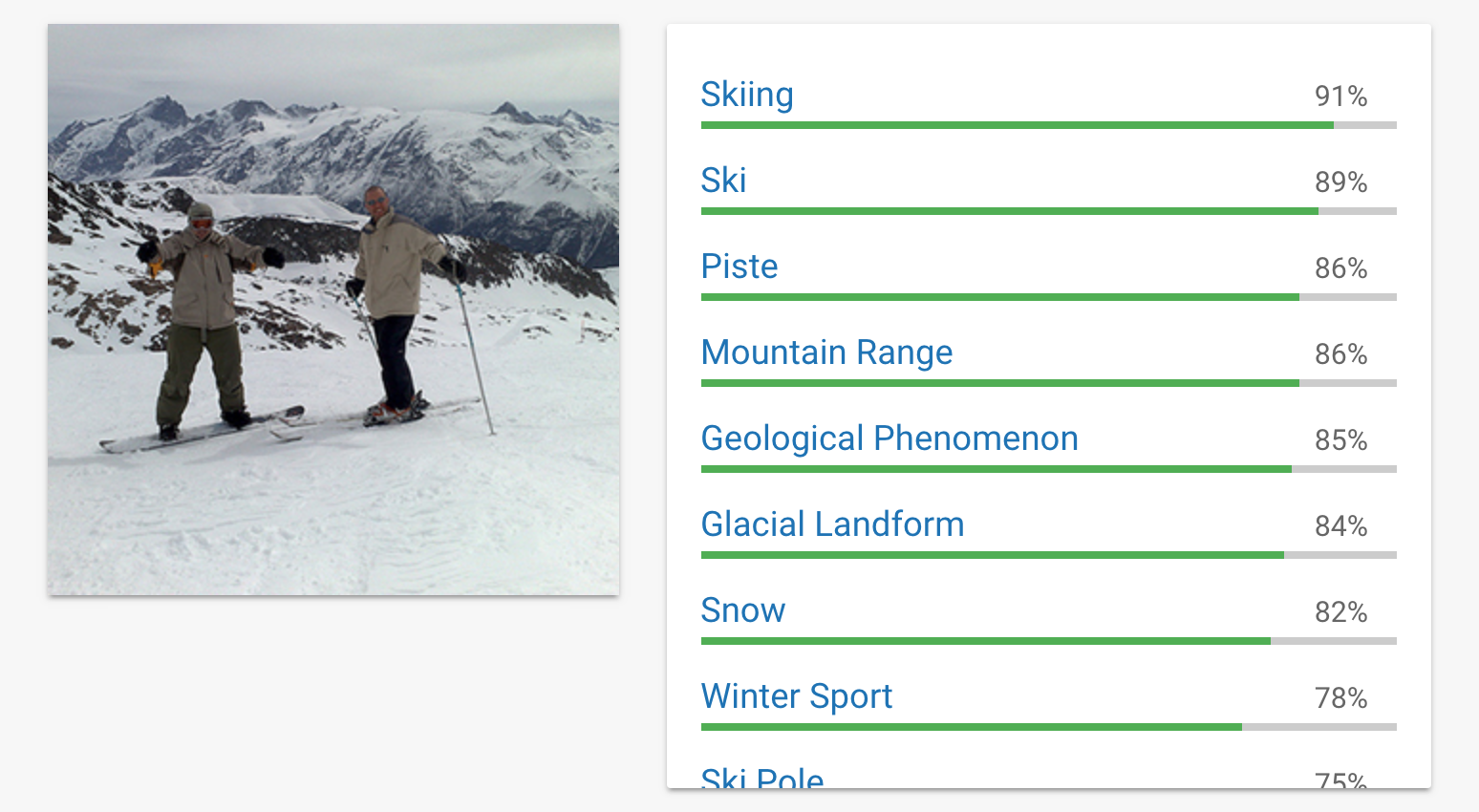}
    \caption{The Google Cloud Vision Demo labeling on the unperturbed image.}
    \label{fig:gcv-targeted}
    \includegraphics[width=0.8\linewidth]{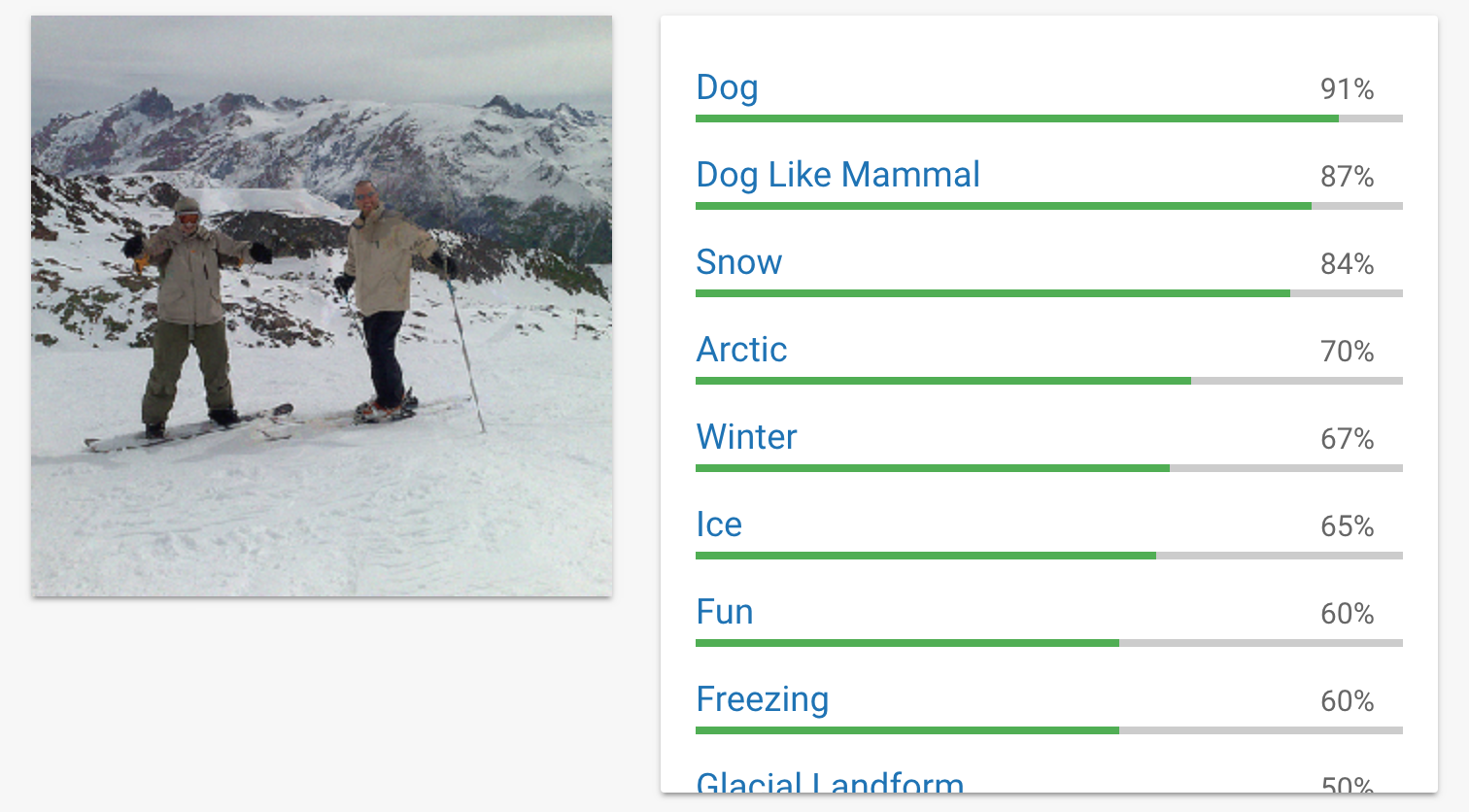}
    \caption{The Google Cloud Vision Demo labeling on the adversarial image
	generated with $\ell_\infty$ bounded perturbation with $\epsilon = 0.1$:
	the image is labeled as the target class.}
    \label{fig:gcv-targeted-adv}
\end{figure}

Figure~\ref{fig:gcv-targeted} shows an unperturbed image being correctly
labeled as several skiing-related classes, including ``skiing'' and ``ski.'' We
run our partial-information attack to force this image to be classified as
``dog'' (an arbitrarily chosen target class). Note that the label ``dog'' does not appear in the output for the
unperturbed image. Using our partial-information algorithm, we initialize our
attack with a photograph of a dog (classified by GCV as a dog) and successfully
synthesize an image that looks like the skiers but is classified as ``dog,'' as
shown in Figure~\ref{fig:gcv-targeted-adv}~\footnote{\url{https://www.youtube.com/watch?v=1h9bU7WBTUg}
demonstrates our algorithm transforming the image of a dog into an image of the
skier while retaining the original classification}.

\section{Related work}
\label{sec:related-work}

\citet{biggio} and \citet{szegedy} discovered that machine learning classifiers
are vulnerable to adversarial examples. Since then,
a number of techniques have been developed to generate
adversarial examples in the white-box
case~\cite{iclr2015-goodfellow,sp2017-carlini, cvpr2016-dezfooli, dezfooli-universal, hayes-universal},
where an attacker has full access to the model parameters and architecture.

In this section, we focus on prior work that specifically address the black-box case and practical
attack settings more generally and compare them to our contributions.
Throughout this section, it is useful to keep in the mind the axes for comparison:
(1) white-box vs. black-box; (2) access to train-time
information + query access vs. only query access; (3) the scale of the targeted model
and the dataset it was trained on (MNIST vs. CIFAR-10 vs. ImageNet);
(4) untargeted vs. targeted.

\subsection{Black-box adversarial attacks}

Several papers have investigated practical black-box attacks on real-world
systems such as speech recognition systems~\cite{hiddenvoice}, malware
detectors~\cite{hu-malware,xu-malware}, and face recognition
systems~\cite{sharif}. Current black-box attacks use either substitute networks
or gradient estimation techniques.

\subsubsection{Black-box attacks with substitute networks}

One approach to generating adversarial examples in the black-box case is with a
substitute model, where an adversary trains a new model with synthesized data
labeled by using the target model as an oracle. Adversarial
examples can then be generated for the substitute with white-box methods, and they will
often transfer to the target model, even if it has a different architecture or training dataset~
\cite{szegedy,iclr2015-goodfellow}. \citet{papernot16, papernot17}
have successfully used this method to attack commercial classifiers like
the Google Cloud Prediction API, the Amazon Web Services Oracle, and the MetaMind API,
even evading various defenses against adversarial attacks.
A notable subtlety is that the Google Cloud Vision API~\footnote{https://cloud.google.com/vision/}
we attack in this work is \textit{not} the same as the Google Cloud Prediction API~\footnote{https://cloud.google.com/prediction/docs/} (now
the Google Cloud Machine Learning Engine) attacked in \citet{papernot16, papernot17}. Both systems
are black-box, but the Prediction API is intended to be trained with the user's
own data, while the Cloud Vision API has been trained on large amounts of Google's own
data and works ``out-of-the-box.'' In the black-box threat model we consider in
our work, the adversary does not have access to the internals of the model
architecture and has no knowledge of how the model was trained or what datasets
were used.

\citet{papernot16, papernot17} trained the Cloud Prediction API with small
datasets like MNIST and successfully demonstrated an untargeted attack.
As~\citet{iclr2017-liu} demonstrated, it is more difficult to transfer targeted adversarial
examples with or without their target labels, particularly when attacking
models trained on large datasets like ImageNet. Using ensemble-based methods,
\citet{iclr2017-liu} overcame these limitations to attack the Clarifai API.
Their threat model specifies that the adversary does not have any
knowledge of the targeted model, its training process, or training and
testing data, matching our definition of black-box. While Liu et al.'s
substitute network attack does not require any queries to the target model (the
models in the ensemble are all trained on ImageNet), only 18\% of the targeted
adversarial examples generated by the ensemble model are transferable in the
Clarifai attack. In contrast, our method needs to query the model many times to perform a
similar attack but has better guarantees that an adversarial example will
be generated successfully (94\% even in the partial-information case, and
over 99\% in the standard black-box setting).

\subsubsection{Black-box attacks with gradient estimation}

\citet{zoo} explore black-box gradient estimation methods as an alternative to
substitute networks, where we have noted that transferability is not always
reliable. They work under the same threat model, restricting an adversary solely to
querying the target model as an oracle. As they note, applying zeroth order optimization
naively in this case is not a tractable solution, requiring $2 \times 299 \times 299 \times 3 = 536406$
queries to estimate the gradients with respect to all pixels.
To resolve this problem, they devise an iterative coordinate descent procedure
to decrease the number of evaluations needed and successfully
perform untargeted and targeted attacks on MNIST and CIFAR-10 and untargeted attacks on
ImageNet. Although we do not provide a direct comparison due to the
incompability of the $\ell_2$ and $\ell_\infty$ metric as well as the
fixed-budget nature of the optimization algorithm in~\citet{zoo}, our
method takes far fewer queries to generate imperceptible adversarial
examples.

\citet{narodytska} propose a black-box gradient estimation attack using a local-search
based technique, showing that perturbing only a small fraction of pixels
in an image is often sufficient for it to be misclassified. They successfully perform
targeted black-box attacks on an ImageNet classifier with only query access and additionally with a
more constrained threat model where an adversary only has access to
a ``proxy'' model. For the most successful misclassification attack on
CIFAR-10 (70\% success) the method takes 17,000 queries on average.
Targeted adversarial attacks on ImageNet are not considered.

\subsection{Adversarial attacks with limited information}

Our work is concurrent with~\citet{brendel}, which also explores the label-only case using their ``Boundary
Attack,'' which is similar to our two-step partial information algorithm. Starting with an image of the target
adversarial class, they alternate between taking steps on the decision boundary to maintain the adversarial
classification of the image and taking steps towards the original image.

\subsection{Other adversarial attacks}

Several notable works in adversarial examples use similar techniques but with different adversarial goals or threat models.
\citet{xu-malware} explore black-box adversarial examples to fool PDF malware classifiers. To generate
an adversarial PDF, they start with an instance of a malicious PDF and use genetic algorithms to evolve
it into a PDF that is classified as benign but preserves its malicious behavior. This attack is similar
in spirit to our partial-information algorithm, although our technique (NES) is more similar to traditional
gradient-based techniques than evolutionary algorithms, and we consider multiway image classifiers
under a wider set of threat models rather than binary classifiers for PDFs. \citet{nguyen} is another
work that uses genetic algorithms and gradient ascent to produce images that fool
a classifier, but their adversarial goal is different: instead of aiming to make a interpretable image of
some class (e.g. skiiers) be misclassified as another class (e.g. a dog), they generate entirely unrecognizable
images of noise or abstract patterns that are classified as a paricular class. Another work generates
adversarial examples by inverting the image instead of taking local steps; their goal is to show that
CNNs do not generalize to inverted images, rather than to demonstrate a novel attack or to consider
a new threat model~\cite{hosseini}.

\section{Conclusion}
\label{sec:conclusion}

Our work defines three new black-box threat models that characterize many real
world systems: the query-limited setting, partial-information
setting, and the label-only setting. We introduce new algorithms for
attacking classifiers under each of these threat models and show the
effectiveness of these algorithms by attacking an ImageNet classifier. Finally, we
demonstrate targeted adversarial examples for the Google Cloud Vision API,
showing that our methods enable black-box attacks on real-world systems in
challenging settings. Our results suggest that machine learning systems remain
vulnerable even with limited queries and information.

\ifdefined\isaccepted
    \section*{Acknowledgements}

We wish to thank Nat Friedman and Daniel Gross for providing compute resources
for this work.

\fi
{\small
\bibliography{paper}

\begin{thebibliography}{31}
\providecommand{\natexlab}[1]{#1}
\providecommand{\url}[1]{\texttt{#1}}
\expandafter\ifx\csname urlstyle\endcsname\relax
  \providecommand{\doi}[1]{doi: #1}\else
  \providecommand{\doi}{doi: \begingroup \urlstyle{rm}\Url}\fi

\bibitem[Athalye et~al.(2017)Athalye, Engstrom, Ilyas, and Kwok]{robustadv}
Athalye, A., Engstrom, L., Ilyas, A., and Kwok, K.
\newblock Synthesizing robust adversarial examples.
\newblock 2017.
\newblock URL \url{https://arxiv.org/abs/1707.07397}.

\bibitem[Biggio et~al.(2012)Biggio, Nelson, and Laskov]{biggio}
Biggio, B., Nelson, B., and Laskov, P.
\newblock Poisoning attacks against support vector machines.
\newblock In \emph{Proceedings of the 29th International Coference on
  International Conference on Machine Learning}, ICML'12, pp.\  1467--1474,
  2012.
\newblock ISBN 978-1-4503-1285-1.
\newblock URL \url{http://dl.acm.org/citation.cfm?id=3042573.3042761}.

\bibitem[Biggio et~al.(2013)Biggio, Corona, Maiorca, Nelson, {\v{S}}rndi{\'c},
  Laskov, Giacinto, and Roli]{biggio2013evasion}
Biggio, B., Corona, I., Maiorca, D., Nelson, B., {\v{S}}rndi{\'c}, N., Laskov,
  P., Giacinto, G., and Roli, F.
\newblock Evasion attacks against machine learning at test time.
\newblock In \emph{Joint European Conference on Machine Learning and Knowledge
  Discovery in Databases}, pp.\  387--402. Springer, 2013.

\bibitem[Brendel et~al.(2018)Brendel, Rauber, and Bethge]{brendel}
Brendel, W., Rauber, J., and Bethge, M.
\newblock Decision-based adversarial attacks: Reliable attacks against
  black-box machine learning models.
\newblock In \emph{Proceedings of the International Conference on Learning
  Representations (ICLR)}, 2018.
\newblock URL \url{https://arxiv.org/abs/1712.04248}.

\bibitem[Carlini \& Wagner(2017)Carlini and Wagner]{sp2017-carlini}
Carlini, N. and Wagner, D.
\newblock Towards evaluating the robustness of neural networks.
\newblock In \emph{IEEE Symposium on Security \& Privacy}, 2017.

\bibitem[Carlini et~al.(2016)Carlini, Mishra, Vaidya, Zhang, Sherr, Shields,
  Wagner, and Zhou]{hiddenvoice}
Carlini, N., Mishra, P., Vaidya, T., Zhang, Y., Sherr, M., Shields, C., Wagner,
  D., and Zhou, W.
\newblock Hidden voice commands.
\newblock In \emph{25th USENIX Security Symposium (USENIX Security 16), Austin,
  TX}, 2016.

\bibitem[Chen et~al.(2017)Chen, Zhang, Sharma, Yi, and Hsieh]{zoo}
Chen, P.-Y., Zhang, H., Sharma, Y., Yi, J., and Hsieh, C.-J.
\newblock Zoo: Zeroth order optimization based black-box attacks to deep neural
  networks without training substitute models.
\newblock In \emph{Proceedings of the 10th ACM Workshop on Artificial
  Intelligence and Security}, AISec '17, pp.\  15--26, New York, NY, USA, 2017.
  ACM.
\newblock ISBN 978-1-4503-5202-4.
\newblock \doi{10.1145/3128572.3140448}.
\newblock URL \url{http://doi.acm.org/10.1145/3128572.3140448}.

\bibitem[Dasgupta et~al.(2006)Dasgupta, Hsu, and Verma]{mixedgaussian}
Dasgupta, S., Hsu, D., and Verma, N.
\newblock A concentration theorem for projections.
\newblock In \emph{Conference on Uncertainty in Artificial Intelligence}, 2006.

\bibitem[Evtimov et~al.(2017)Evtimov, Eykholt, Fernandes, Kohno, Li, Prakash,
  Rahmati, and Song]{stopsign}
Evtimov, I., Eykholt, K., Fernandes, E., Kohno, T., Li, B., Prakash, A.,
  Rahmati, A., and Song, D.
\newblock Robust physical-world attacks on machine learning models.
\newblock \emph{CoRR}, abs/1707.08945, 2017.

\bibitem[Goodfellow et~al.(2015)Goodfellow, Shlens, and
  Szegedy]{iclr2015-goodfellow}
Goodfellow, I.~J., Shlens, J., and Szegedy, C.
\newblock Explaining and harnessing adversarial examples.
\newblock In \emph{Proceedings of the International Conference on Learning
  Representations (ICLR)}, 2015.

\bibitem[Gorban et~al.(2016)Gorban, Tyukin, Prokhorov, and
  Sofeikov]{quasiortho}
Gorban, A.~N., Tyukin, I.~Y., Prokhorov, D.~V., and Sofeikov, K.~I.
\newblock Approximation with random bases.
\newblock \emph{Inf. Sci.}, 364\penalty0 (C):\penalty0 129--145, October 2016.
\newblock ISSN 0020-0255.
\newblock \doi{10.1016/j.ins.2015.09.021}.
\newblock URL \url{http://dx.doi.org/10.1016/j.ins.2015.09.021}.

\bibitem[Hayes \& Danezis(2017)Hayes and Danezis]{hayes-universal}
Hayes, J. and Danezis, G.
\newblock Machine learning as an adversarial service: Learning black-box
  adversarial examples.
\newblock \emph{CoRR}, abs/1708.05207, 2017.

\bibitem[Hosseini et~al.(2017)Hosseini, Xiao, Jaiswal, and
  Poovendran]{hosseini}
Hosseini, H., Xiao, B., Jaiswal, M., and Poovendran, R.
\newblock On the limitation of convolutional neural networks in recognizing
  negative images.
\newblock \emph{2017 16th IEEE International Conference on Machine Learning and
  Applications (ICMLA)}, pp.\  352--358, 2017.

\bibitem[Hu \& Tan(2017)Hu and Tan]{hu-malware}
Hu, W. and Tan, Y.
\newblock Black-box attacks against {RNN} based malware detection algorithms.
\newblock \emph{CoRR}, abs/1705.08131, 2017.

\bibitem[Kurakin et~al.(2016)Kurakin, Goodfellow, and
  Bengio]{goodfellow-physical}
Kurakin, A., Goodfellow, I., and Bengio, S.
\newblock Adversarial examples in the physical world.
\newblock 2016.
\newblock URL \url{https://arxiv.org/abs/1607.02533}.

\bibitem[Liu et~al.(2017)Liu, Chen, Liu, and Song]{iclr2017-liu}
Liu, Y., Chen, X., Liu, C., and Song, D.
\newblock Delving into transferable adversarial examples and black-box attacks.
\newblock In \emph{Proceedings of the International Conference on Learning
  Representations (ICLR)}, 2017.

\bibitem[Madry et~al.(2017)Madry, Makelov, Schmidt, Tsipras, and
  Vladu]{madry-adversarial}
Madry, A., Makelov, A., Schmidt, L., Tsipras, D., and Vladu, A.
\newblock Towards deep learning models resistant to adversarial attacks.
\newblock 2017.
\newblock URL \url{https://arxiv.org/abs/1706.06083}.

\bibitem[Moosavi{-}Dezfooli et~al.(2017)Moosavi{-}Dezfooli, Fawzi, Fawzi, and
  Frossard]{dezfooli-universal}
Moosavi{-}Dezfooli, S., Fawzi, A., Fawzi, O., and Frossard, P.
\newblock Universal adversarial perturbations.
\newblock In \emph{{CVPR}}, pp.\  86--94. {IEEE} Computer Society, 2017.

\bibitem[Moosavi-Dezfooli et~al.(2016)Moosavi-Dezfooli, Fawzi, and
  Frossard]{cvpr2016-dezfooli}
Moosavi-Dezfooli, S.-M., Fawzi, A., and Frossard, P.
\newblock Deepfool: a simple and accurate method to fool deep neural networks.
\newblock In \emph{IEEE Conference on Computer Vision and Pattern Recognition
  (CVPR)}, 2016.

\bibitem[Narodytska \& Kasiviswanathan(2017)Narodytska and
  Kasiviswanathan]{narodytska}
Narodytska, N. and Kasiviswanathan, S.~P.
\newblock Simple black-box adversarial perturbations for deep networks.
\newblock In \emph{IEEE Conference on Computer Vision and Pattern Recognition
  (CVPR)}, 2017.

\bibitem[Nesterov \& Spokoiny(2017)Nesterov and Spokoiny]{nesterov}
Nesterov, Y. and Spokoiny, V.
\newblock Random gradient-free minimization of convex functions.
\newblock \emph{Found. Comput. Math.}, 17\penalty0 (2):\penalty0 527--566,
  April 2017.
\newblock ISSN 1615-3375.
\newblock \doi{10.1007/s10208-015-9296-2}.
\newblock URL \url{https://doi.org/10.1007/s10208-015-9296-2}.

\bibitem[Nguyen et~al.(2014)Nguyen, Yosinski, and Clune]{nguyen}
Nguyen, A.~M., Yosinski, J., and Clune, J.
\newblock Deep neural networks are easily fooled: High confidence predictions
  for unrecognizable images.
\newblock \emph{CoRR}, abs/1412.1897, 2014.

\bibitem[Papernot et~al.(2016{\natexlab{a}})Papernot, McDaniel, and
  Goodfellow]{papernot16}
Papernot, N., McDaniel, P., and Goodfellow, I.
\newblock Transferability in machine learning: from phenomena to black-box
  attacks using adversarial samples.
\newblock 2016{\natexlab{a}}.

\bibitem[Papernot et~al.(2016{\natexlab{b}})Papernot, McDaniel, Jha,
  Fredrikson, Celik, and Swami]{sp2016:papernot}
Papernot, N., McDaniel, P., Jha, S., Fredrikson, M., Celik, Z.~B., and Swami,
  A.
\newblock The limitations of deep learning in adversarial settings.
\newblock In \emph{IEEE European Symposium on Security \& Privacy},
  2016{\natexlab{b}}.

\bibitem[Papernot et~al.(2017)Papernot, McDaniel, Goodfellow, Jha, Celik, and
  Swami]{papernot17}
Papernot, N., McDaniel, P., Goodfellow, I., Jha, S., Celik, Z.~B., and Swami,
  A.
\newblock Practical black-box attacks against machine learning.
\newblock In \emph{Proceedings of the 2017 ACM on Asia Conference on Computer
  and Communications Security}, ASIA CCS '17, pp.\  506--519, New York, NY,
  USA, 2017. ACM.
\newblock ISBN 978-1-4503-4944-4.
\newblock \doi{10.1145/3052973.3053009}.
\newblock URL \url{http://doi.acm.org/10.1145/3052973.3053009}.

\bibitem[Salimans et~al.(2017)Salimans, Ho, Chen, and Sutskever]{salimans}
Salimans, T., Ho, J., Chen, X., and Sutskever, I.
\newblock Evolution strategies as a scalable alternative to reinforcement
  learning.
\newblock \emph{CoRR}, abs/1703.03864, 2017.
\newblock URL \url{http://arxiv.org/abs/1703.03864}.

\bibitem[Sharif et~al.(2017)Sharif, Bhagavatula, Bauer, and Reiter]{sharif}
Sharif, M., Bhagavatula, S., Bauer, L., and Reiter, M.~K.
\newblock Adversarial generative nets: Neural network attacks on
  state-of-the-art face recognition.
\newblock 2017.

\bibitem[Szegedy et~al.(2013)Szegedy, Zaremba, Sutskever, Bruna, Erhan,
  Goodfellow, and Fergus]{szegedy}
Szegedy, C., Zaremba, W., Sutskever, I., Bruna, J., Erhan, D., Goodfellow, I.,
  and Fergus, R.
\newblock Intriguing properties of neural networks.
\newblock 2013.
\newblock URL \url{https://arxiv.org/abs/1312.6199}.

\bibitem[Szegedy et~al.(2015)Szegedy, Vanhoucke, Ioffe, Shlens, and
  Wojna]{szegedy-inception}
Szegedy, C., Vanhoucke, V., Ioffe, S., Shlens, J., and Wojna, Z.
\newblock Rethinking the inception architecture for computer vision.
\newblock 2015.
\newblock URL \url{https://arxiv.org/abs/1512.00567}.

\bibitem[Wierstra et~al.(2014)Wierstra, Schaul, Glasmachers, Sun, Peters, and
  Schmidhuber]{nes}
Wierstra, D., Schaul, T., Glasmachers, T., Sun, Y., Peters, J., and
  Schmidhuber, J.
\newblock Natural evolution strategies.
\newblock \emph{J. Mach. Learn. Res.}, 15\penalty0 (1):\penalty0 949--980,
  January 2014.
\newblock ISSN 1532-4435.
\newblock URL \url{http://dl.acm.org/citation.cfm?id=2627435.2638566}.

\bibitem[Xu et~al.(2016)Xu, Yi, and Evans]{xu-malware}
Xu, W., Yi, Y., and Evans, D.
\newblock Automatically evading classifiers: A case study on pdf malware
  classifiers.
\newblock \emph{Network and Distributed System Security Symposium (NDSS)},
  2016.

\end{thebibliography}
\bibliographystyle{icml2018}
}

\end{document}